\documentclass[10pt,twocolumn,letterpaper]{article}

\usepackage[pagenumbers]{cvpr} 

\usepackage[dvipsnames]{xcolor}

\definecolor{cvprblue}{rgb}{0.21,0.49,0.74}
\usepackage[pagebackref,breaklinks,colorlinks,citecolor=cvprblue]{hyperref}

\def\paperID{8} 
\def\confName{CVPR}
\def\confYear{2024}

\title{uTRAND: Unsupervised Anomaly Detection in Traffic Trajectories}

\author{Giacomo D'Amicantonio\\
Eindhoven University of Technology\\
Eindhoven, Netherlands\\
{\tt\small g.d.amicantonio@tue.nl}
\and
Egor Bondarau\\
Eindhoven University of Technology\\
Eindhoven, Netherlands\\
{\tt\small e.bondarev@tue.nl}
\and 
Peter H.N de With\\
Eindhoven University of Technology\\
Eindhoven, Netherlands\\
{\tt\small p.h.n.de.with@tue.nl}
}

\begin{document}
\maketitle
\begin{abstract}
    Deep learning-based approaches have achieved significant improvements on public video anomaly datasets, but often do not perform well in real-world applications. This paper addresses two issues: the lack of labeled data and the difficulty of explaining the predictions of a neural network. To this end, we present a framework called uTRAND, that shifts the problem of anomalous trajectory prediction from the pixel space to a semantic-topological domain. The framework detects and tracks all types of traffic agents in bird's-eye-view videos of traffic cameras mounted at an intersection. By conceptualizing the intersection as a patch-based graph, it is shown that the framework learns and models the normal behaviour of traffic agents without costly manual labeling. Furthermore, uTRAND allows to formulate simple rules to classify anomalous trajectories in a way suited for human interpretation. We show that uTRAND outperforms other state-of-the-art approaches on a dataset of anomalous trajectories collected in a real-world setting, while producing explainable detection results.
\end{abstract}
\section{Introduction}
\label{sec:intro}
In the domain of traffic surveillance, the task of anomaly detection has been subject to a progressive evolution. Initially, attempts have been made to devise expert systems characterized by complex sets of rules, aiming at emulating the intricacies of traffic dynamics. However, the implementation and scalability of such systems are impeded by the inherent complexity and variability of traffic behavior, often necessitating a disproportional number of exceptions to accommodate diverse scenarios.
Recent advancements in computer vision are bolstered by the proliferation of sensor data from traffic monitoring infrastructure, which allow for a paradigm shift in anomaly detection methodologies. These modern approaches leverage the computational capabilities of neural networks to analyze data streams obtained from traffic cameras. While exhibiting promising performance on standardized benchmarks, these methodologies have concurrently shown critical aspects such as privacy preservation, algorithmic biases, and false-positive mitigation. 
\begin{figure*}[t]
\begin{center}
\includegraphics[width=1\textwidth]{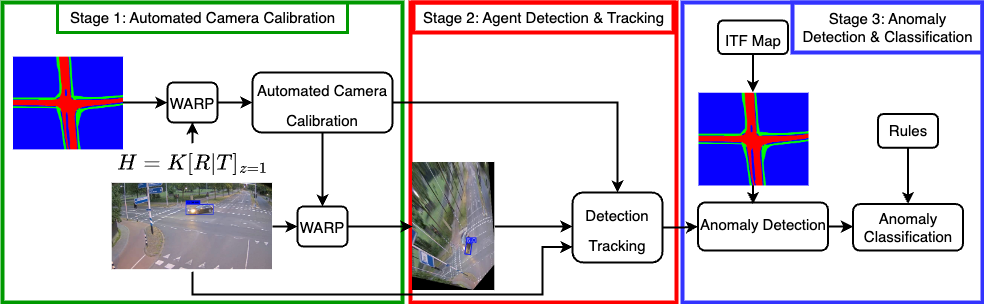}
\end{center}
   \caption{Overview of uTRAND. At the first stage, the cameras are calibrated and the videos are warped, obtaining their BEV videos. At the second stage, the traffic agents are detected and tracked in the camera video and translated to the BEV video. In the last stage, the intersection is split in semantic patches (nodes). The previously detected agents are associated with the patch they occupy at each frame. The links between the nodes and normal behavior of an agent in that node are learned by analyzing a few hours of camera videos through the framework, modeling the dynamics of the intersection. An agent that deviates from the modeled behaviors is detected as anomalous. The anomaly can further be classified by using simple rules that do not require domain knowledge.}
\label{fig:overview}
\end{figure*}

This paper presents a methodological framework designed to address the aforementioned challenges in anomaly detection within traffic intersections called uTRAND (Unsupervised TRajectory ANomaly Detector). Leveraging neural network architectures, the proposed approach detects and tracks moving traffic agents, by translating raw sensor data into a structured semantic-topological space. This framework allows to formulate simple anomaly detection rules. To validate the efficacy of uTRAND in practical scenarios, we experiment on a real-world intersection environment, leveraging a three-camera setup to capture anomalous traffic behavior. Comparative evaluations against existing approaches demonstrate the superior performance of the proposed framework in terms of anomaly detection accuracy and false-positive mitigation, while adhering to stringent deployment criteria delineated in this work.
\section{Related Work}
The field of traffic anomaly detection has seen significant contributions, categorized in three main groups: unsupervised (Section~2.1), weakly supervised (Section~2.2), and supervised (Section~2.3) approaches. 
\subsection{Unsupervised Anomaly Detection}
Unsupervised methods rely on the assumption that any trajectory diverging from regular patterns indicates an anomaly. Common approaches to the anomaly detection task~\cite{chandola, sophia, santhosh} define the problem as a clustering task, employing high-dimensional representations to distinguish regular and anomalous trajectories. Despite varied implementations and conceptual approaches, these methods collectively encounter challenges related to computational complexity. Notably, D'Acierno et al.~\cite{acierno} have addressed these computational issues. Such methodologies find their applications in both anomaly detection and classification tasks.

Concerning unsupervised deep learning methodologies, the Variational Auto-Encoder (VAE) proposed by Kingma et al.~\cite{kingma, kignma2} is a prominent model for anomaly detection. Approaches based on VAEs leverage unsupervised training to acquire latent representations of dataset distributions, thereby enabling the identification of anomalies by detecting deviations from learned representations~\cite{lin, chen, zhou, niu}. While some studies have applied these techniques to traffic video analysis~\cite{roy, santosh2, kumaran}, privacy-related concerns arise due to the utilization of visual features. More recently, this has resulted in efforts to address such issues~\cite{dong, guang, liu, luo}. Another avenue of research focuses on frame reconstruction within videos, where anomalies are detected based on discrepancies between generated and original frames~\cite{ionescu, chen, dong}. However, these methods may fail when considering subtle differences in trajectories.
\subsection{Weakly-Supervised Anomaly Detection}
In weakly-supervised anomaly detection methodologies, the task of identifying anomalies within video data often adopts a Multiple Instance Learning (MIL) framework. Each video instance is annotated with a label indicating the presence of anomalies, while models are trained to predict the onset and duration of these anomalies, which ensures accurate classification based on both aspects. This approach has been extensively explored in the literature, as evidenced by studies~\cite{hu, wu, tian, zhang}. Despite leveraging the temporal dimension of videos in various ways, these methods are frequently constrained by limitations in visual representations. A common challenge encountered in such approaches is the discrepancy between training and deployment scenes, wherein background pixels may hamper model performance and generalization, even when not directly implied from the detected anomalous behavior.

Recent advancements have witnessed the application of transformer architectures to this domain. For instance, in~\cite{li}, attention mechanisms are employed to capture correlations between video-level and snippet-level anomalies. Similarly,~\cite{chen2} proposes to utilize transformer networks to mitigate scene-inconsistencies, which demonstrates higher efficacy on prevalent public datasets.
\begin{figure*}[ht]
\begin{center}
\includegraphics[width=0.95\textwidth]{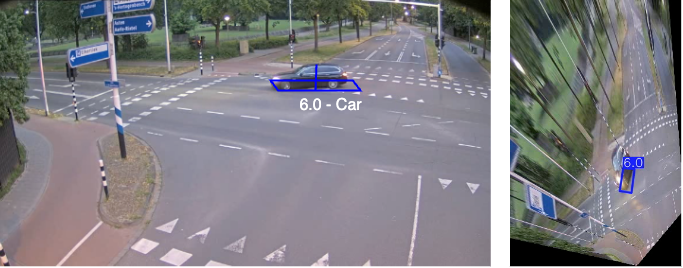}
\end{center}
   \caption{The agents are initially detected in the camera view. YoloV8 estimates a bounding box for each agent, assigns to it an ID, and tracks it across frames. Subsequently, the detected bounding boxes are projected in the BEV, where the three dimensional bounding boxes are detected, carrying over the id assigned to the agent in the camera view.}
\label{fig:detection}
\end{figure*}
\subsection{Supervised Anomaly Detection}
Supervised anomaly detection frameworks typically involve training classifier models with anomaly-level labels. A common class of methods in this field are algorithms such as Support Vector Machines (SVM)~\cite{batapati, piciarelli}, operating on the premise of discerning anomalous trajectories from regular ones. However, these methods often lack the required flexibility to effectively handle diverse anomaly types, such as discrepancies in trajectory length.  While some approaches adopt direct supervision~\cite{sultani, sarker}, others explore semi-supervised or self-supervised learning strategies~\cite{wu2, georgescu, huang}. However, the acquisition and labeling of a sufficiently large dataset for effective model training can be prohibitively expensive. Furthermore, the labeled classes in the data define a specific set of classes which are detected by the model, while the others require a general out-of-dataset class or a dedicated fine-tuning to be detected. This aspect is relevant to control the occurrence of false positives. 

A significant limitation of neural networks is their inherent lack of explainability~\cite{nguyen, wu2}. Recently, hybrid frameworks have emerged to address this issue. For instance, in~\cite{aboah, minni}, neural networks are utilized for pattern and feature extraction from video data, while anomaly detection is performed using more insightful methods like decision trees. The framework introduced in this paper aligns with this hybrid paradigm, offering both feature extraction capabilities and enhanced explainability, while preserving the privacy of citizens when detecting an anomaly.
\section{Methodology}
The proposed trajectory-based anomaly detection framework uTRAND is designed to meet the following criteria.

\begin{itemize}
    \item \textit{Human-interpretable predictions}: the system is designed to produce predictions that are comprehensible to human users.
    \item \textit{Reduced dependency on labeled data}: the framework operates autonomously without necessitating extensive collections of labeled video data, facilitating its deployment in real-world conditions.
\end{itemize}

\begin{figure*}[t]
\begin{center}
\includegraphics[width=0.45\textwidth]{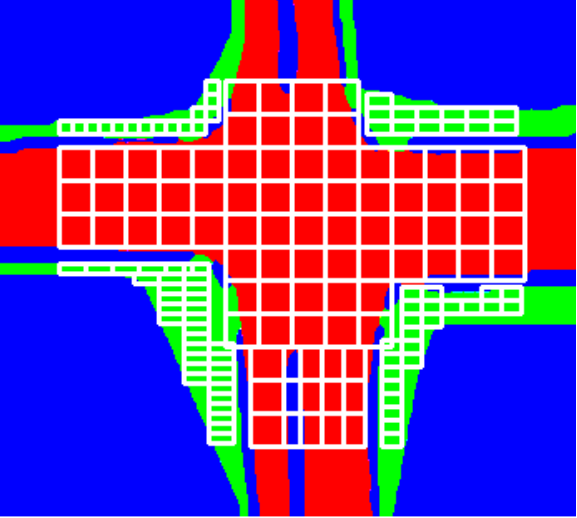}
\includegraphics[width=0.45\textwidth]{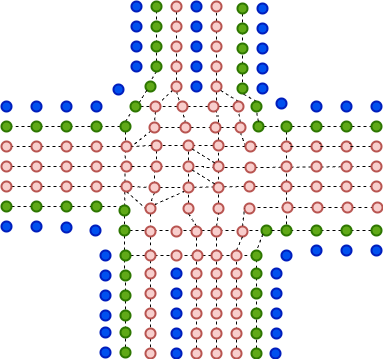}
\end{center}
   \caption{The intersection is split into patches according to the ITF map. The nodes can be of four different classes: road, bicycle lane, curb and crosswalk. The framework establishes connections between the nodes in which traffic agents move more often.}
\label{fig:topview}
\end{figure*}
To address the above criteria, we shift the detection of anomalies in trajectories from the camera video domain to a semantic graph representation derived from the bird's-eye-view (BEV) of the intersection. The framework learns regular behavioral patterns of agents traversing intersections through unsupervised analysis of unlabeled data, thereby enabling the detection of anomalous trajectories. Moreover, this approach facilitates the formulation of straightforward rules for classifying different anomaly types.
The architecture of uTRAND is depicted in Figure~\ref{fig:overview} and comprises of three primary components: automated camera calibration, agent detection and tracking, and anomaly detection.
\subsection{Automated Camera Calibration}
The initial step in the framework involves the automated calibration of cameras located within the intersection. After segmenting the BEV representation of the intersection in three semantic regions (road, bicycle path, and walkable terrain) as depicted in Figure~\ref{fig:topview}, The calibration is performed by a Graph Neural Network (GNN)-based technique, which is described in~\cite{mine}. For the sequel of this paper, we will refer to the semantically segmented BEV of the intersection as the BEV intersection. To facilitate this segmentation, intrinsic and extrinsic parameters of a Pan-Tilt-Zoom (PTZ) camera model are sampled. The BEV is then warped using the homography transformations obtained from the sampled parameters, to generate synthetic images of different viewpoints of the same intersection. 

The GNN is trained on these semantically-segmented synthetic images to predict the homography matrices of each camera, efficiently mapping pixels from the camera views to the BEV. The homography projection operates under the assumption that every pixel lies on a planar surface, which is valid within the semantic domain and does not compromise the method's generality. Leveraging the estimated homographies, the camera videos are warped to generate the BEV representations of each camera field-of-view, thereby enabling comprehensive analysis and anomaly detection within the intersection environment.
\subsection{Agent Detection and Tracking}
The second component detects and tracks traffic agents in the camera videos and projects their bounding boxes in the BEV videos obtained by warping the camera video, as shown in Figure~\ref{fig:detection}. The agents detected and tracked are vehicles, pedestrians and bicyclists. To this end, the approach proposed by~\cite{3dbb} is updated to use the more efficient YoloV8 model~\cite{yolo}. Each agent detected in the camera videos is assigned a unique ID and is tracked across frames. The model leverages the estimated homography to warp the feature maps of the camera video produced by different layers of the model. These feature maps are concatenated with the feature maps obtained from the BEV video. 
As a result, the model produces a bounding box for each agent in the BEV video, representing the base of the 3D bounding box, and an orientation angle indicating the direction of movement. At the end of this stage, the agents are detected, tracked and localized in the BEV of the intersection at each frame.
\begin{figure}[ht]
\begin{center}
\includegraphics[width=0.47\textwidth]{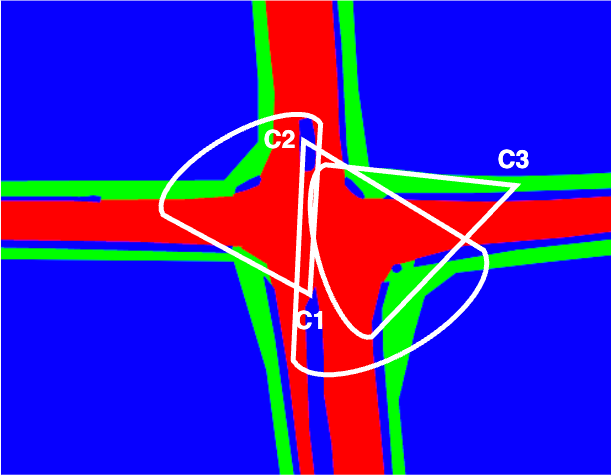}
\end{center}
   \caption{Semantically segmented bird's-eye-view of the intersection (called BEV intersection). The fields-of-view of the three cameras overlap at the center of the intersection.}
\label{fig:topviewfov}
\end{figure}
\begin{figure*}
\centering
\begin{subfigure}[b]{0.41\textwidth} 
    \includegraphics[width=\linewidth]{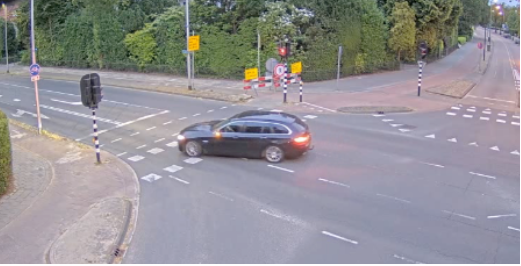}
    \caption{Camera view of an \textit{improper turn} anomaly.} 
    \label{fig:wrongturnimg}
\end{subfigure}%
\quad  
\begin{subfigure}[b]{0.41\textwidth} 
    \includegraphics[width=\linewidth]{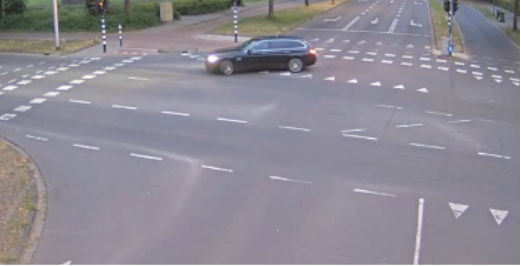}
    \caption{Camera view of an \textit{improper turn (donut)} anomaly.} 
    \label{fig:donutimg}
\end{subfigure}%
\quad
\begin{subfigure}[b]{0.41\textwidth} 
    \includegraphics[width=\linewidth]{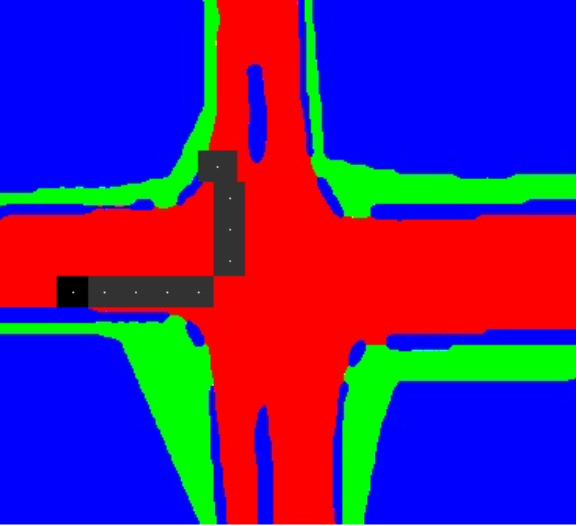}
    \caption{Anomalous trajectory.} 
    \label{fig:wrongturntraj}
\end{subfigure}%
\quad
\begin{subfigure}[b]{0.41\textwidth} 
    \centering
    \includegraphics[width=\linewidth]{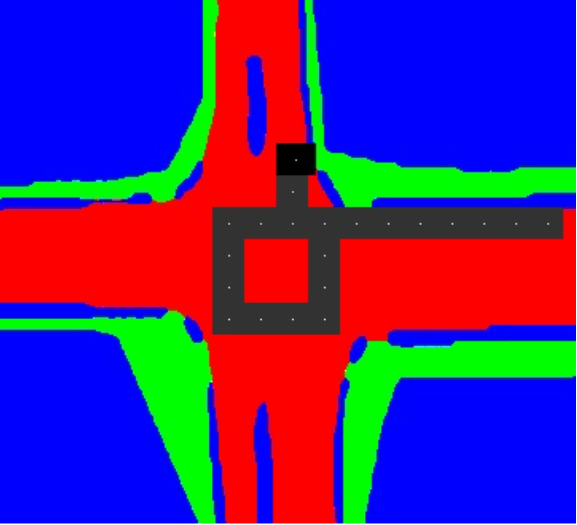}
    \caption{Anomalous trajectory.} 
    \label{fig:donutraj}
\end{subfigure}
\caption{Camera views on the same intersection and the corresponding trajectory visualization. Top row: two frames of a vehicle performing improper turns. In subfigure~\ref{fig:wrongturnimg}, the car is turning right and then drives in the opposite lane, while in subfigure~\ref{fig:donutimg} the same vehicle is driving in a circle in the intersection. Bottom row: the uTRAND's visualization of the trajectories of the two actions. Both trajectories are correctly detected and classified as anomalies.}  
\label{fig:comparison}
\end{figure*}
\subsection{Anomaly Detection}
In the final stage of the framework, the BEV intersection undergoes an automated partitioning into patches utilizing the Intersection Topology Format (ITF) maps~\cite{itf}, as shown in Figure~\ref{fig:topview}. The ITF map models the topological layout of the intersection, delineating features such as ingress and exgress points of road lanes and bicycle paths, expressed in world coordinates. It is also possible to perform an intersection partitioning into patches by utilizing the segmentation masks of the road markings~\cite{vijverberg2024}. Leveraging the camera calibration parameters acquired in the initial stage, these world coordinates are accurately projected onto the BEV videos, thereby facilitating the mapping of world coordinates to pixels. 
Subsequently, these pixels are utilized to automatically partition the semantic regions of the intersection into discrete patches. Each patch of the BEV intersection corresponds to a segment of a lane, bicycle path, or curb. Patches associated with crosswalks constitute an additional type of a patch. Agent association with patches is achieved by calculating the Intersection over Union (IoU) between the BEV bounding box of an agent and the patch itself. These patches are conceptualized as nodes within a graph representing the entire intersection. 
The relationships between nodes are learned through observation of agent movements within the intersection, establishing links between nodes, as an agent moves from one to another. This enables the framework to learn and model the agents behaviors within the intersection, characterized by three attributes for each node:\\

\begin{itemize}
\setlength\itemsep{1em}
    \item $S_{node_i} = \{node^{t+1}_k, ... , node^{t+1}_n\}$ the set of nodes to which an agent normally proceeds at (video) frame time $t+1$ given $node_i$ at time $t$,
    \item $T_{avg}$ is average time spent in a node,
    \item $A$ is a set that contains the types of agents allowed in the node.
\end{itemize}

These attributes are learned by processing several hours of camera videos, using the described pipeline. As illustrated in Figure~\ref{fig:topviewfov}, there is an overlap in the field of view (FOV) of the cameras. Consequently, if the cameras are synchronized, it becomes feasible to track an agent's trajectory, as it transitions between the FOVs of different cameras. This capability enables the comprehensive analysis of an agent's complete trajectory within the intersection.
In this context, if the trajectory of an agent deviates from the learned regular behavioral model, the trajectory is automatically identified as anomalous in real time. 
In conclusion, the framework facilitates the formulation of rules that categorize anomalous behavior in simple terms using the previously defined attributes.
For example, an unlawful turn by an agent from $node_1$ to $node_2$ at time $t$ can be classified by the rule:
\begin{equation}
     node_2 \notin S_{node_1}.
\end{equation}
Furthermore, an agent in an unlawful position, e.g. a bicycle on the road at $node_k$, can be classified in the following way:
\begin{equation}
    bicycle \notin A_{node_k},
\end{equation}
where $A_{node_k}$ is the set of agent types allowed in $node_k$. Finally, an agent that stays in $node_l$ from $t_i$ to $t_n$ can be classified as unlawfully stopping via the rule:
\begin{equation}
      \mathbf{card}\{t_i, ..., t_n\} > T^{node_l}_{avg}.
\end{equation} 
The previous examples provide a set of rules to classify three types of anomalous trajectories. The formulation of such rules defines the framework to detect and classify anomalous trajectories of agents without explicitly labeling videos of anomalous and normal trajectories. Consequently, uTRAND is part of the unsupervised learning paradigm.

\section{Experiments}
The proposed framework is evaluated on video data collected at a traffic intersection, of which the BEV intersection is shown in Figure~\ref{fig:topviewfov}. The BEV intersection presents bicycle paths and curbs on the side of every road. Three cameras were mounted on light poles and cover the majority of the intersection. 
The model is set to learn the attributes of each node within the intersection, by observing the trajectories of traffic agents during a 12 hour temporal interval, covering the afternoon and evening from 12 pm to 12 am. This interval encompasses a diverse range of typical activities, including vehicular, cyclist and pedestrian moving in the intersection.
\begin{table}[ht]
\caption{Amount of anomalous actions recorded. The anomalies are split in three classes: \textit{agent in improper zone} (i.e. pedestrians and bicyclists on the road, vehicle parked on the curb), \textit{unlawful stop} (i.e. car stopping at the center of the intersection) and \textit{improper turn} (vehicles turning in the wrong direction, changing lanes, driving in a circle or doing U-turns).}
\centering
\begin{tabular}{l|c}
    \toprule
    \textbf{Anomaly} & \textbf{Amount} \\
    \midrule
    Agent in improper zone & 24\\
    Unlawful stop & 3 \\
    Improper turn & 14 \\    
    \midrule
    \textbf{Total} & \textbf{41}\\
    \bottomrule
\end{tabular} 
\label{tab:anomalyoverview}
\end{table} 
\begin{table}[t]
\caption{Comparison between uTRAND and previous works. The baseline methods have been chosen based on their performances on public datasets. The $F_1$~score is the chosen comparison metric.}
\centering
\begin{tabular}{l|c|c}
    \toprule
    \textbf{Method} & \textbf{Pre-training} & \textit{$\mathbf{F_1(\%)}$} \\
    \midrule
    MONAD\cite{monad} & UCSD Ped2\cite{ucsd} & 62.35\\
    HF2VAD\cite{hf2vad} & ShanghaiTech\cite{shangai} & 69.62 \\
    AI-VAD\cite{aivad} & ShanghaiTech\cite{shangai} & 72.49 \\
    \midrule
    \textbf{uTRAND} & - & \textbf{82.89} \\
    \bottomrule
\end{tabular} 
\label{tab:comparison}
\end{table} 
\begin{table}[t]
\caption{Performance of uTRAND on the classification task. The accuracy of the framework is reported along with the False Positive rates (FP) for each class.}
\centering
\begin{tabular}{l|c|c}
    \toprule
    \textbf{Anomaly} & \textbf{Accuracy (\%)} & \textbf{FP} \\
    \midrule
    Improper turn & 71.43 & 0.46\\
    Unlawful stop & 66.67 & 0.28 \\
    Agent in improper zone & 91.16 & 0.54\\    
    \bottomrule
\end{tabular} 
\label{tab:classification}
\end{table} 
Each node is connected to all neighboring nodes, meaning that each node is connected to all adjacent patches in the intersection. The inter-node links are uniformly initialized with unity weights. After this initialization, each traversal of an agent between two nodes contributes to a unity increment to the corresponding link's weight. Upon observation and learning of trajectories within the 12-hour dataset, links characterized by a weight falling below an empirically determined threshold are pruned. This empirically established threshold is essential for mitigating localization inaccuracies, encountered during the second stage of the framework. Specifically, in crowded scenes the orientation of the estimated bounding boxes can change across consecutive frames while the direction of movement of an agent remains the same, resulting in an erroneous assignment during the third stage of the framework.

To evaluate the efficacy of the framework, various anomalous behaviors have been authorized, played and recorded. The dataset comprising anomalous trajectories contains 41~instances, as detailed in Table~\ref{tab:anomalyoverview}. The performance of uTRAND is compared with three anomaly detection methods in Table~\ref{tab:comparison}. The selected benchmark methods are also based on the unsupervised learning paradigm, as mentioned in Section~2. 

The uTRAND framework outperforms other approaches by a significant $F_1$~score margin, showing a notable performance improvement. Additionally, anomalous trajectories are visually depicted, as illustrated in Figure~\ref{fig:comparison}. The visualization produced by the framework portrays the logic of the anomaly detection, therefore it is intuitive for a human to understand why a trajectory is detected as anomalous. This is an important advantage over other methods that are based on estimation of an anomalous score computed by a neural network. In fact, such scores can be influenced by a significant amount of spurious correlations that the model may infer from a training dataset, thereby reducing the interpretation capability of the results. 

Moreover, with the rules expressed in Section~3.3, the uTRAND framework exhibits the capability to classify anomalous actions. The performance of the proposed method on the classification task is shown in Table~\ref{tab:classification}. It is important to note that the instances of false positive identifications are heavily dependent upon the precision of the camera calibration and agent detection stages. 
The straightforward formulation of the classification rules permits the extension of the framework's final stage to classify additional types of anomalous trajectories neither by re-training the model from scratch nor employing any transfer learning and fine-tuning techniques. This capability of the framework represents a significant advantage over methods relying on neural networks to classify actions.

\section{Conclusion}
This study proposes a three-stage framework called uTRAND, for detecting and classifying abnormal traffic trajectories within an intersection. The framework leverages on the intersection's topological layout by transposing the problem domain onto a semantic graph structure. We demonstrate that within this domain, accurate detection of abnormal trajectories is achievable, yielding interpretable predictions. The classification of anomalies relies on straightforward principles, rendering the framework adaptive to real-world scenarios and being interpretable for end-users.
Further exploration in this field could benefit from more accurate object detection and tracking algorithms. While uTRAND demonstrates efficacy in detecting and classifying single-agent anomalous trajectories, additional efforts are necessary to tailor this approach towards multi-agent trajectory analysis. One possible research direction entails the application of neuro-symbolic or causality-based methodologies to address this challenge.
{
    \small
    \bibliographystyle{ieeenat_fullname}
    \bibliography{main}
}

\end{document}